# Learning Sign Language Representation using CNN-LSTM, 3DCNN, CNN–RNN–LSTM and CCN-TD

Nikita Louison
Department of Computing and
Information Technology
The University of the West Indies
St. Augustine Campus
nikita.louison@my.uwi.edu

Wayne Goodrige
Department of Computing and
Information Technology
The University of the West Indies
St. Augustine Campus
wayne.goodridge@sta.uwi.edu

Koffka Khan
Department of Computing and
Information Technology
The University of the West Indies
St. Augustine Campus
koffka.khan@sta.uwi.edu

*Abstract*— Existing Sign Language Learning applications focus on the demonstration of the sign in the hope that the student will copy a sign correctly. In these cases, only a teacher can confirm that the sign was completed correctly, by reviewing a video captured manually. Sign Language Translation is a widely explored field in visual recognition. This paper seeks to explore the algorithms that will allow for real-time, video sign translation, and grading of sign language accuracy for new sign language users. This required algorithms capable of recognizing and processing spatial and temporal features. The aim of this paper is to evaluate and identify the best neural network algorithm that can facilitate a sign language tuition system of this nature. Modern popular algorithms including CNN and 3DCNN are compared on a dataset not yet explored, Trinidad and Tobago Sign Language as well as an American Sign Language dataset. The 3DCNN algorithm was found to be the best performing neural network algorithm from these systems with 91% accuracy in the TTSL dataset and 83% accuracy in the ASL dataset.

## I. Introduction

Sign languages are languages that use visual-manual movement to convey meaning. Sign languages are expressed through manual articulations in combination with non-manual elements. Sign languages are legitimate and mature natural languages with their own lexicon and grammar that can also be specific to regions and nations. In Trinidad and Tobago, the indigenous deaf sign language is referred to as Trinidad and Tobago Sign Language (TTSL) [8], occasionally called Trinidadian or Trinbago Sign Language (TSL). ASL or American Sign Language is also taught in this country and the wider region [25].

Sign Language serves as the main communication tool in the deaf community, as well as between hearing and deaf people if the former is learning to sign. It conveys information and meaning by way of spatio-temporal visual patterns, which are created by manual (handshapes) [42] and non-manual cues (facial expressions and upper body motion) [22]. When the first deaf school opened in Trinidad and Tobago in 1943 this birthed the Trinidad and Tobago Sign Language (TTSL) [9]. The establishment of the school provided a path of development for a national deaf community and a unique sign language. There is a need to bridge the gap between sign language users or the hearing impaired, according to the Director of the Ministry's Disability Affairs Unit in Trinidad and Tobago, most persons who are mute, deaf, or hard of hearing, depend on the availability and willingness of a family member or close friend to translate.

Introducing an educational application for hearing individuals to learn and practice basic signs and phrases with correction and "real-time" feedback will be an aid to sign language education in the community. This would involve the uploading of videos and/or real-time capture of students practicing signing for translation and assessment. The challenge is to find the best method for sign language recognition (SLR) [1] and thereby classification, as these videos will include both temporal and spatial features; with the context of each frame being relative to the previous, and each frame holding important information [29]. This is needed in SLR where the meaning of a sign is derived from the subtle differences in the combination of head poses, body and manual motions, that translate into various meanings.

Neural Networks (NN) true and modern-day rise came in 2010 and has kept growing up to the present day. The shift in relevance came when major improvements were discovered for data processing. Graphical Processing Units (GPU) [45] were found to be the best tools for faster data processing in NN. In addition, techniques were developed in NN to adapt and improve performance over Support Vector Machines (SVM) [20], including its hyperparameter tunings such as epoch, activation functions, batch normalization, optimizer algorithms and pre-trained layers for transfer learning. While searching for a solution to Sign Language Translation (SLT) [27] and Sign Language Recognition (SLR) [1] neural networks are the most discussed classification solution. Some networks encountered are Convolutional Neural Networks (CNN) [40], Recurrent Neural Network (RNN), Long short-term memory (LSTM) [41] and Three Dimensional Convolutional Neural Networks (3DCNN) [12].

Recognition of sign videos as well as computer-based processing is also linked to gesture recognition [2]. While significant progress has been made in this field a lot of the research done was noted to heavily be in the human action recognition field [43]. In the cases of SL NN studies, some of the language datasets include ASL (American) [7], ISL (Indian) [37] and KSL (Korean) [19]. More study is then required in the field of SLR and NN with use of feasible or accessible technologies focusing on sign languages used in the Caribbean region. However, the focus of our work is to identify and compare modern and popular Deep Neural Network models and to determine as well as test the best one that facilitates Sign Language Translation for Trinidad and Tobago Sign Language (TTSL) and American Sign Language (ASL) dataset.

The research objective is to implement a NN solution that allows for the use of smart devices, and web or laptop cameras to capture video for Sign language recognition (SLR). The goal is to identify the best NN model and/or combination that facilitates SLR and SLT in TTSL, a dataset not previously explored. Choosing neural networks as the method to analyze and read sign language is best, as NN are known to adapt to changing input, as such the network generates the best possible result without needing to redesign the output criteria. NNs are also better positioned to handle image processing than formerly more prominent deep learning methods such as SVM because of technology such as GPU and even Tensor Processing Unit (TPU) [26] that are geared towards increasing speed of processing in NNs.

What performance with respect to accuracy can be obtained for SLT & SLR using deep learning and TTSL and ASL video data, and which combination of different models will achieve the best score in terms of performance and accuracy? Will the combination of CNN-RNN-LSTM [18] continue to stand above other 'state of the art' algorithms such as 3DCNN in terms of accuracy and performance? Once the best is discovered it can be implemented/tested in a learning application that corrects or confirms correct sign language forms. We will identify the optimal combination of different NN models for accuracy and performance considering the preconditions, in particular the data as well as the time scope and resources available. We also find room for improvement and further research. In addition, what is being investigated is the best performing NN model when trained, saved and used in a prediction scenario.

This paper consists of six Sections. In Section 2 we introduce the four models, CNN-LSTM [16], 3DCNN [12], CNN-RNN-LSTM [18] and CNN-TD. Section III describes the tools, data collection techniques and experimental steps. The Results and Discussion is given in Sections IV and V. Finally, in Section VI the conclusion and future work is stated.

II. METHODS

This section will describe the proposed solutions to SLR in terms of the neural network algorithms and the formulae that make up the algorithms. The interaction of layers in each NN algorithm is also worked out, as well as the method employed for their combination.

*A. CNN-LSTM Architecture*

The CNN Long Short-Term Memory Network, or CNN LSTM, is an LSTM architecture intended primarily for sequence prediction tasks using spatial inputs such as pictures or videos. Consider a model with a CNN on the input, LSTM in the middle, and MLP at the output that employs a stack of layers [46]. This type of model can scan a series of picture inputs, such as a video, and make a forecast. The CNN-LSTM Model Layer breakdown used by us is given in Figure 1.

The CNN-LSTM model used in this paper is the Convnet LSTM Model. It is based on the ConvnetLSTM2D Model and is a prebuilt Keras Model [31] used for the CNN-LSTM layer application [32]. This layer of the network is followed by the decision network which facilitates the classification and output.

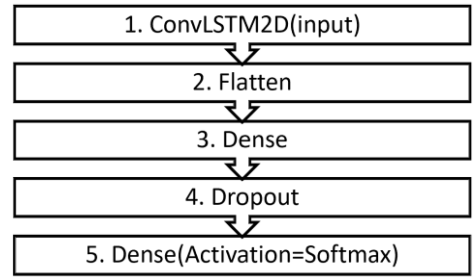

Fig. 1. CNN-LSTM Model Layer breakdown.

*B. 3DCNN*

The 3D Convolution layer consists of a set of learnable 3D filters, each of which has a small local receptive field that extends across all input channels [3]. A nonlinear function rectified linear unit (ReLU) activation [24], this function has a simple byproduct which is to speed up network training. Every 3D kernel in the first layer is convolved to a volume of the input stacking frames to produce a spatiotemporal feature map. The 3D kernels in the consecutive layers are likewise convolved to a volume of stacking feature maps produced by the preceding layers. The 3DCNN Model Layer breakdown used by us is given in Figure 2.

The model used in this paper is based on [39]. It should be noted that the total parameters are the highest among all the models, see Figure 2. The 3DCNN will be tested with feature learning, by extracting the local spatiotemporal features of sign sequences. Google's TensorFlow with Keras' Conv3D model will be used to execute the 3D Convolutional layers for feature learning.

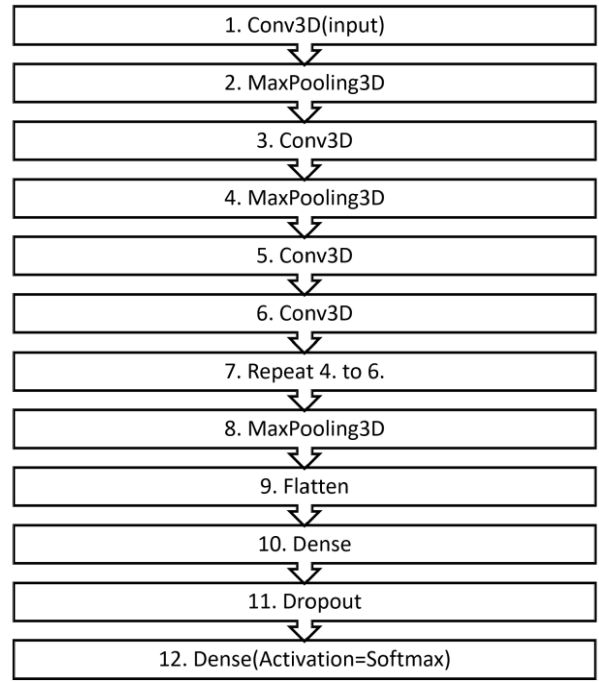

Fig. 2. 3DCNN Model Layer breakdown.

*C. CNN-RNN-LSTM*

The architecture assessed by us is the CNN-RNN-LSTM Model and it is based on the model mentioned in [14]. It differs from the CNN-LSTM Architecture as in CNN-LSTM the learning combination is completely prebuilt, whereas in CNN-RNN-LSTM the layers are all defined individually.

The CNN-RNN-LSTM Model Layer breakdown used by us is given in Figure 3.

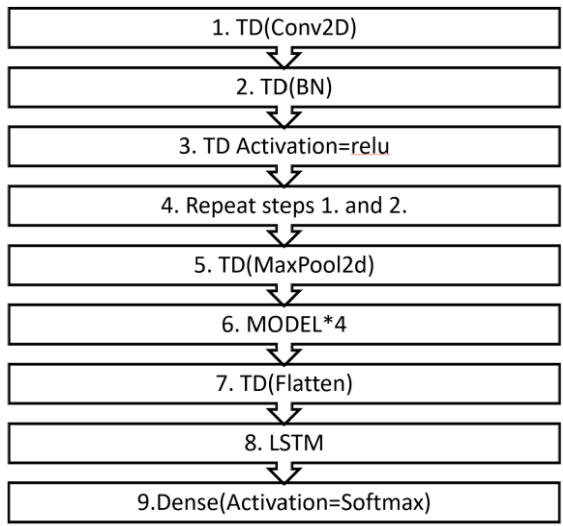

Fig. 3. CNN-RNN-LSTM Model Layer breakdown.

*D. CNN-TD (Time Distributed)*

An addition that can be made to the CNN approach is to apply a Time Distributed layer or wrapper around the convolutional and Pooling Layers [35] and [33]. Time-Distributed layer applies the same layer to several batch inputs, and it produces one output per input to get the result "in time". It is a way CNN can work with Images in sequence on its own with or without LSTM implementation to improve it. The CNN-TD Model Layer breakdown used by us is given in Figure 4.

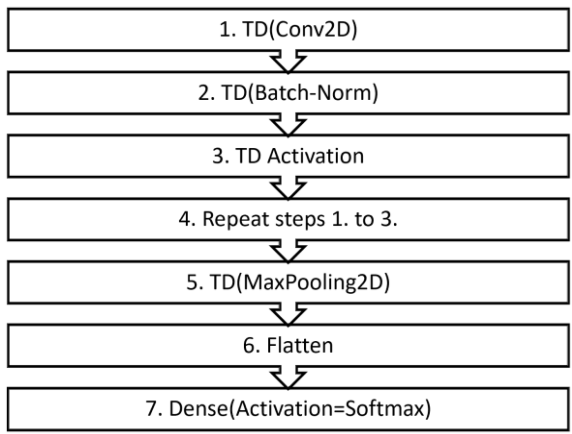

Fig. 4. CNN-TD Model Layer breakdown.

### III. EXPERIMENTS

In this Section we discuss the following topics. The dataset from two sources is first described. Further, we discuss the tools used to run the experiments. These tools will be used to perform a series of experiments on combined TTSL and ASL datasets implemented in python 3 using Keras and TensorFlow backend as the deep-learning framework. After we present how our research compares models (see Section II) in terms of pre-processing, feature extraction and classification. Then we discuss the experimental steps and finally assumptions and limitations.

*A. Datasets*

The dataset used in this work includes TTSL as well as ASL. In the literature review, we will find that sign language translation systems have been researched for ASL, ISL and even KSL. Nothing before exists regarding research for TTSL. This research will help evaluate the performance of these current and modern approaches on the dataset. Hopefully, this paper can present an experiment on the model identified as "most ideal" on TTSL videos or video streams and while promoting learning the indigenous language sign language of Trinidad and Tobago.

*B. Tools*

TensorFlow (TF) is an open-source deep learning framework that the Google Brain [34] team developed in 2012 and released in 2015. It possesses readily available APIs that can save time and speeds up the process of training a model. Keras is one such deep learning application programming interface (API), written in the python language [6], and designed for fast experimentation in deep neural networks is used for conducting tests in this paper. TensorFlow [49] also offers native systems for model deployment to production such as TF Serve and TF Lite, the first which is beneficial for future use of the model in an application. Another similar framework is Pytorch [21], whose deployment systems are very new, and its documentation is low in comparison to TF. In this research paper, Keras and TensorFlow are utilized and run on the Google Colab [10] service. An additional benefit is a high scalability where it is possible to run code either on CPU, GPU, or across a cluster of these systems for training.

While TF may be the framework chosen, some pre-built models will be accessed on this system for testing. One such model is the Inception Network [4], a type of CNN initially known as GoogLeNet. It was made by Google in 2014 and released in 2015 and has been constantly enhanced. The Inception model is an intricate network of layers, organized in modules that act as smaller models inside the larger model [38]. Individual modules have many convulsions with each result stacking together, allowing for the elimination of the need to decide beforehand what type of convolutions to perform, making the model more general. The third update of this system is utilized in this research paper.

In order to test CNN-LSTM, a prebuilt Keras Model was selected for implementation called ConvLSTM2D. The CNN to LSTM implementation has all the steps or layers built in so that this model can be applied as a layer in the network. It is used before implementing the decision network Dense layer that precedes the Softmax activation, which produces the results from classification.

To test the effectiveness of using a CNN model on its own the "Time Distributed Wrapping" class is used. This class can be used with CNN and even LSTM and allows the application of a layer to every temporal slice of an input. Essentially the wrapper can be applied to a Conv2D input layer, that accepts the batch input, to each timestep identified independently. Time Distributed applies the same instance of Conv2D to each of the timestamps, where the same set of weights will be used at each timestamp.

For 3CDNN processing, Google's TensorFlow with Keras' Conv3D model can be utilized. It must be partnered with the MaxPooling3D model to down-sample the feature maps or making them smaller, without losing information which saves computational resources.

Finally, as previously mentioned the TensorFlow models will be run on The Google Colab Service. It is a Python development environment that runs in any web browser using Google Cloud. Colaboratory, or "Colab", is a Google Research product that allows the writing and executing of python code on a web browser. It is a hosted Jupyter notebook service that provides access to computing resources such as CPUs or GPUs.

*C. Data Collection, Video Input and Pre-processing*

The proposed approaches in this paper are evaluated by a data set created and produced by the research team. It is composed of ten sign words, six of which are ASL and four are TTSL. For each sign, there are between 10 to 20 videos each. This data was collected by an application written using python programming utilizing OpenCV. This application captured the sign being demonstrated, saved a video copy of the sign and then split the videos into frames. These frames were captured in black and white as well as color. The video as well as the frames captured were numbered, labelled, and stored in folders corresponding to the name of the sign.

The dataset which is referring to the captured frames was divided by training, validation or testing, with training being 80% and testing and validation 20%. Details of the dataset are depicted in a separate section and sample action categories are given in Figure 5 and Table I respectively.

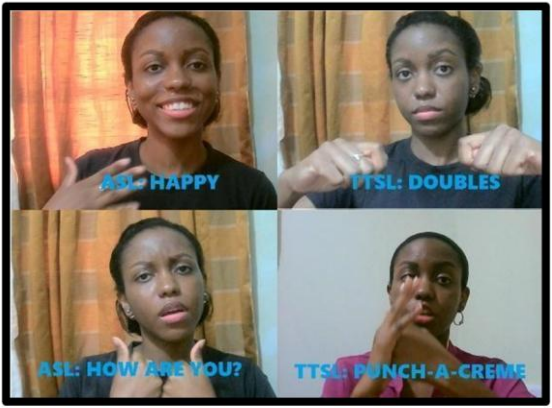

Fig. 5. Samples of actions from ASL & TTSL Datasets

TABLE I. THE SIGNS USED FOR TRAINING. FOUR SIGNS THAT ARE TTSL ARE DOUBLES, PARANG, SORREL AND PUNCH-A-CRÈME

| Sign | Bye | Happy | Doubles | Hello | Help | How are you? | Thanks | Parang | Punch-a-Creme | Sorrel |
|---|---|---|---|---|---|---|---|---|---|---|
| Number of Videos | 20 | 20 | 20 | 20 | 20 | 20 | 20 | 10 | 10 | 10 |

The method used to capture the video and split it into frames can be used for real data capture and analysis for assessing the accuracy of signs once the best-case model is identified. Also involved in preprocessing is the organizing of the video datasets for easy access from Google Colab. Each video that was previously created and organized by sign was placed into folders stored in the Google drive cloud service. The name of the folders again corresponds with the sign and doubles as a label. These are the classes the NNs will learn to predict when the models are trained.

*D. Experiment Steps*

The videos were broken up by the TTSL and ASL datasets, then the frames were extracted, reshaped and stored in memory temporarily. Then each dataset was parsed through each model. The Classes of the ASL Dataset are the signs for "Bye," "Happy," "Hello," "Help," "How are you" and "Thanks." The Classes of the TTSL dataset are signs: "Doubles," "Parang," "Punch-a-crème" and "Sorrel."

Prebuilt models were selected and applied to relevant layers in the model setup. While the models used for training differed, as previously mentioned the loss function and optimizer used was consistent. Categorical Cross-Entropy also called Softmax Loss, computes the cross-entropy metric between the labels and predictions. Softmax activation is used for multi-class classification and Cross-Entropy for loss. In using this loss function, it will output a probability over the classes (signs) for each image. The Adam optimizer used, is a stochastic gradient descent method that is based on adaptive estimation of first-order and second-order moments [11].

Some hyperparameters [47] used in training were epochs and batch numbers. Epochs being the number of times the learning algorithm will pass through or work on the training dataset. The epochs were either set at a minimum of 15 or a maximum of 20, an early stopping parameter was included to stop the passes if there was a repetition of results five or seven times in a row. Batches are defined as the number of samples to work through at a pass, these were also kept at a smaller number than the sample size. The videos used were limited to a sequence length of thirty-five (35) frames for consistency.

A confusion matrix [48] chart was also produced for each model run where possible. The Confusion matrix is an N x N matrix for appraising the performance of a classification model, where N is the number of target classes. The matrix will compare the actual target values or in this case "target signs" with those predicted by the machine learning model. The Confusion matrices will compare values such as True Positives (TP), False Positives (FP), True Negatives (TN) and False Negatives (FN) to obtain the number of correct and incorrect predictions in a summarized form with count values broken down by each class. An evaluation was also consistently done after training by outputting a classification report on the accuracy, precision, f1 scores, recall and support as well as Training and loss accuracy progression graphs, see Figure 6.

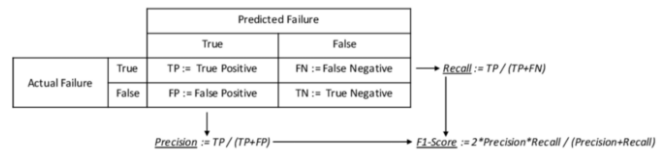

Fig. 6. Confusion matrix, illustrating the calculation of precision, recall, and F1-score. [36]

*E. Assumptions and Limitations*

The entire dataset is not large, with 10 classes or signs with 162 videos created by the researcher. In each class, there are 10 to 20 sample videos of signs under the two main categories of ASL and TTSL. The training, testing and validation will be done on the free tier of Google Colab.

On Table II is shown the "highest" processor features available on the free tier of Colab [17]. Colab allows 12 hours of continuous execution time and provides 12.67 GB RAM and 107.77 GB disk space available for use on their hosted run time. This puts a limit on large datasets that may need to be processed for more than the time limit provided. As the dataset is not large the assumption is that the free or base tier of Colab will be able to handle the processing needed for the experiments planned.

TABLE II. GOOGLE COLAB CPU, GPU AND TPU LIMITS ON THE FREE TIER.

| CPU | GPU | TPU |
|---|---|---|
| Intel Xeon Processor with two cores @ 2.30 GHz and 13 GB RAM | Up to Tesla K80 with 12 GB of GDDR5 VRAM, Intel Xeon Processor with two cores @ 2.20 GHz and 13 GB RAM | Cloud TPU with 180 teraflops of computation, Intel Xeon Processor with two cores @ 2.30 GHz and 13 GB RAM |

While the small datasets may have been an advantage in processing time and resource availability in Google Colab it can have a negative impact on training accuracy. Several workarounds to this were identified. Ensuring that not much down-sampling was occurring during pre-processing specifically, keeping the image size from reducing too small, so that the features that are meant to be captured are still visible. The research was done to identify the best optimizing algorithm and loss function for model training for this multiclass [15] small dataset.

## IV. RESULTS

The performance of each model is discussed by dataset grouping and will be evaluated independently. The dataset along with its challenges are highlighted and evaluation metrics used for assessment are described. The accuracy, recall and efficiency as well as the processes in the models are discussed in more detail. The quantitative and visual results were presented, describing a class-wise performance by precision, recall and F1 scores and summarizing the best and worst-case recognition results of the proposed methods. It should be noted that Sorrel, Punch-a-Crème and Parang signs had the least amount of training videos in the dataset and this impacts recall percentage in the TTSL dataset for all models. A final experiment is done with the highest performing model, and a real-time video.

On Table III shown the number of epochs, patience values and Google Colab Resource usage during training/evaluation of the models.

TABLE III. RUNTIME STATISTICS FOR MODEL TRAINING/EVALUATION

| | #Epochs | Patience | RAM (GB) | GPU(GB) | Learning Rate | Optimizer |
|---|---|---|---|---|---|---|
| CNN-LSTM(ASL) | 13 | 7 | 3.3 | 14.23 | 0.001 | Adam |
| CNN-LSTM(TTSL) | 18 | 7 | 2.77 | 14.23 | 0.001 | Adam |
| 3DCNN(ASL) | 18 | 5 | 2.7 | 14.23 | 0.001 | Adam |
| 3DCNN(TTSL) | 20 | 5 | 2.51 | 8.72 | 0.001 | Adam |
| CNN RNN LSTM (ASL) | 20 | 5 | 3.33 | 14.23 | 0.001 | Adam |
| CNN RNN LSTM (TTSL) | 15 | 5 | 2.76 | 8.71 | 0.001 | Adam |
| CNN(Time Distributed) (ASL) | 17 | 5 | 3.12 | 14.23 | 0.001 | Adam |
| CNN(Time Distributed) (TTSL) | 6 | 5 | 2.88 | 14.24 | 0.001 | Adam |

On all Accuracy and Loss line graphs the Accuracy/Loss is shown on the y-axis and the number of epochs on the x-axis. Categorical Cross-Entropy probability metric was used for loss calculation. Table IV can be used for interpretation of these graphs.

TABLE IV. ACCURACY AND LOSS VALUES INTERPRETION

| | Low Loss | High Loss |
|---|---|---|
| Low Accuracy | A lot of small errors | A lot of big errors |
| High Accuracy | A few small errors | A few big errors |

### A. CNN-LSTM
#### 1) ASL Dataset

The main result of the classification report is shown on Table V. Precision results were an average of 92%, Recall 83% and F1-Score 83%. All signs except for "Bye" and "Thanks" had good recall scores.

TABLE V. CLASSIFICATION REPORT PERCENTAGES BY SIGN FOR THE CNN LSTM MODEL AND ASL DATASET

| "Sign" | Precision | Recall | F1-Score |
|---|---|---|---|
| Bye | 100% | 57% | 73% |
| Happy | 75% | 100% | 86% |
| Hello | 40% | 100% | 57% |
| Help | 100% | 100% | 100% |
| How are you | 100% | 100% | 100% |
| Thanks | 100% | 50% | 67% |

The Accuracy graph displays a relatively smooth ascent with high accuracy values, see Figure 7.

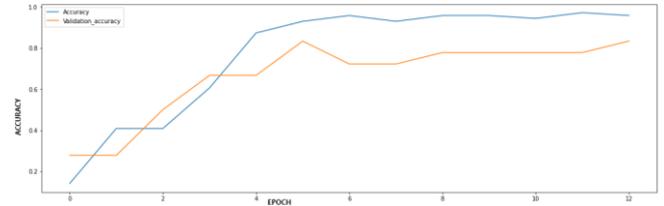

Fig. 7. Accuracy Line Graph Convnet LSTM ASL Dataset

The Loss graph was a smooth descent to completely flat to low loss values indicating few errors when training the dataset, see Figure 8.

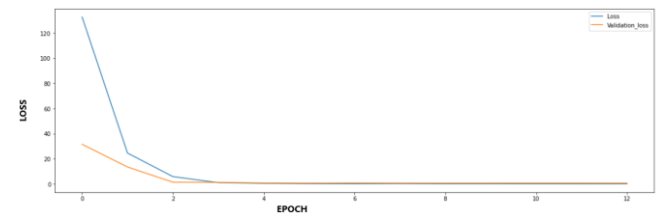

Fig. 8. Loss Chart Convnet LSTM ASL Dataset

Thus, a CNN paired with LSTM produces favorable results for this type of temporal dataset and is supported in the literature.

#### 2) TTSL Dataset

The main result of the classification report is shown on Table VI. Precision results were an average of 81%, Recall 73% and an F1 Score 74%. "Sorrel" had the best recall score and "Punch-a-Crème" the worst.

TABLE VI. CLASSIFICATION REPORT PERCENTAGES BY SIGN FOR THE CNN LSTM MODEL AND TTSL DATASET

| "Sign" | Precision | Recall | F1-Score |
|---|---|---|---|
| Doubles | 100% | 75% | 86% |
| Parang | 60% | 75% | 67% |
| Punch-a-Creme | 100% | 50% | 67% |
| Sorrel | 50% | 100% | 67% |

The Accuracy graph performance here was relatively similar to the ASL dataset displaying a relatively smooth ascent with high accuracy values noted, see Figure 9. The difference is that while accuracy increased the validation accuracy fluctuated but remained high.

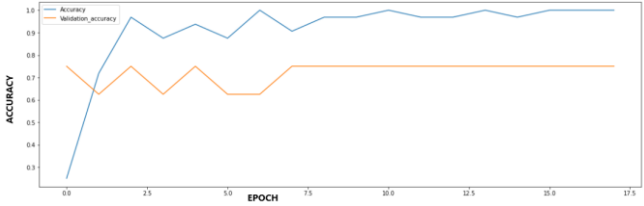

Fig. 9. Accuracy Line Graph Convnet LSTM TTSL Dataset

The Loss graph was a smooth descent from completely flat to low loss values indicating few errors when training the dataset, see Figure 10.

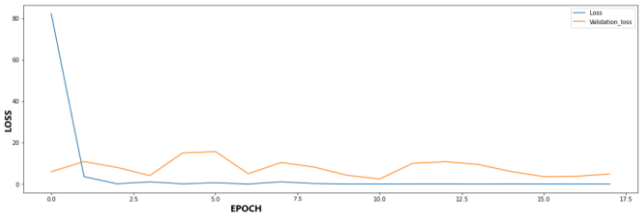

Fig. 10. Loss Line Graph Convnet LSTM TTSL Dataset

The validation loss started low and remained low throughout. This indicates a favorable quality of the training of this model.

*B. 3DCNN*

*a) ASL Dataset:* The main result of the classification report is shown on Table VII. Best sign identification and recall were obtained from signs "Happy", "How are you" and "Thanks".

TABLE VII. CLASSIFICATION REPORT PERCENTAGES BY SIGN FOR THE 3D CNN MODEL AND ASL DATASET

| "Sign" | Precision | Recall | F1-Score |
|---|---|---|---|
| Bye | 100% | 86% | 92% |
| Happy | 100% | 100% | 100% |
| Hello | 50% | 50% | 50% |
| Help | 100% | 50% | 67% |
| How are you | 62% | 100% | 77% |
| Thanks | 100% | 100% | 100% |

The Accuracy graph was not a completely smooth ascension with a dip on the 15th epoch for the accuracy line, see Figure 11.

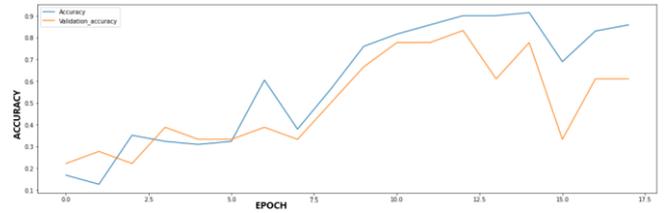

Fig. 11. Accuracy Graph for 3DCNN Model and ASL Dataset

The Loss graph had a straight-line descent by the 2nd Epoch but remained very low for the rest of the passes, see Figure 12.

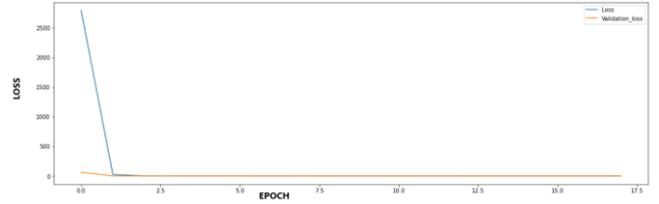

Fig. 12. Loss Graph for 3DCNN Model and ASL Dataset

Overall, the accuracy values remained higher than the errors. These results matched expectations of the 3DCNN Model which thrives on spatio-temporal datasets [44].

*b) TTSL Dataset:* The main result of the classification report is shown on Table VIII. All higher performance values than the ASL dataset. Apart from the other signs, "Punch-a creme" gave the only false positive and had the lowest recall.

TABLE VIII. CLASSIFICATION REPORT RESULTS FOR 3DCNN MODEL AND TTSL DATASET

| "Sign" | Precision | Recall | F1-Score |
|---|---|---|---|
| Doubles | 100% | 100% | 100% |
| Parang | 80% | 100% | 89% |
| Punch-a-Creme | 100% | 50% | 67% |
| Sorrel | 100% | 100% | 100% |

The Accuracy graph was a staggered but continuous ascension by epoch for the accuracy values, see Figure 13.

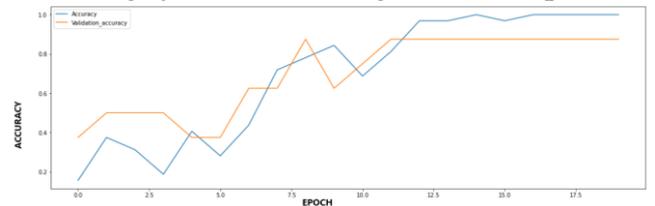

Fig. 13. Accuracy Graph for 3DCNN Model and TTSL Dataset

The Loss graph had a straight-line descent by the 2nd Epoch but remained very low for the rest of the passes, see Figure 14.

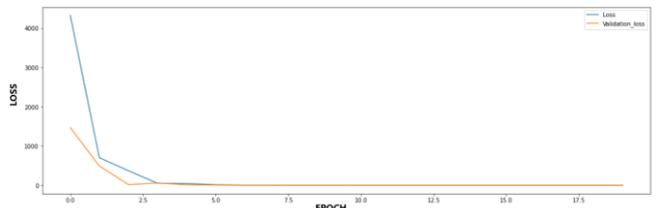

Fig. 14. Loss Graph for 3DCNN Model and TTSL Dataset

Overall, the accuracy values here just as in the ASL dataset remained higher than the errors. Training quality is high.

*C. CNN-RNN-LSTM*

*a) ASL Dataset:* The main result of the classification report is shown on Table IX. Two signs had zero results "Help" and "How are you". The performance on the rest was poor with multiple False positives and very low recall.

TABLE IX. CLASSIFICATION REPORT PERCENTAGES BY SIGN FOR THE CNN-RNN-LSTM MODEL AND ASL DATASET

| "Sign" | Precision | Recall | F1-Score |
|---|---|---|---|
| Bye | 67% | 57% | 62% |
| Happy | 100% | 67% | 80% |
| Hello | 11% | 50 | 18% |
| Help | 0% | 0% | 0% |
| How are you | 0% | 0% | 0% |
| Thanks | 33% | 50% | 40% |

The Accuracy graph displays a slow and small ascension accuracy, see Figure 15.

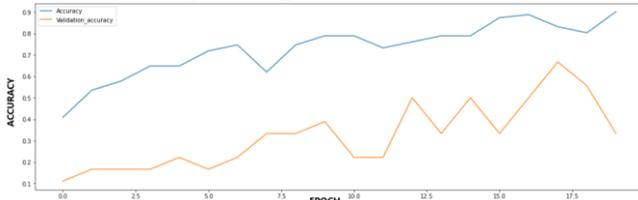
Fig. 15. Accuracy Graph for the CNN-RNN-LSTM Model ASL Dataset

The Loss graph has a better smoother descent however, the loss or error values are significantly higher than the accuracy values which explains the poor performance in the classification report, see Figure 16.

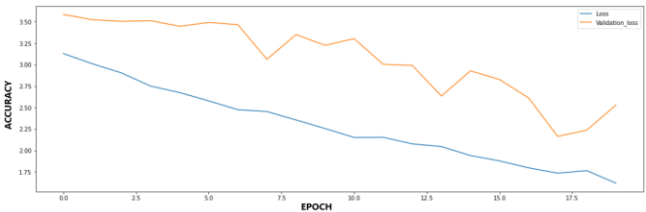
Fig. 16. Loss Graph for the CNN-RNN-LSTM Model ASL Dataset

Training quality was low for this model.

*b) TTSL Dataset:* The main result of the classification report is shown on Table X. The "Punch-a-creme" sign did the worst across all metrics.

TABLE X. CLASSIFICATION REPORT PERCENTAGES BY SIGN FOR THE CNN-RNN-LSTM MODEL TTSL DATASET

| "Sign" | Precision | Recall | F1-Score |
|---|---|---|---|
| Doubles | 100% | 25% | 40% |
| Parang | 40% | 50% | 44% |
| Punch-a-Creme | 0% | 0% | 0% |
| Sorrel | 20% | 100% | 33% |

The Accuracy graph was not a smooth ascension with a significant dip between the 8th and 10th epoch for the validation accuracy, see Figure 17.

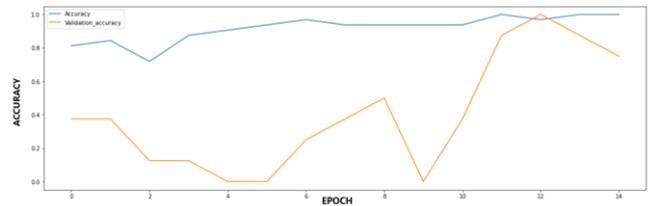
Fig. 17. Accuracy Graph for the CNN-RNN-LSTM Model TTSL Dataset

The Loss graph decreased over epochs, but the values again surpassed the accuracy, meaning the training was of a low quality, see Figure 18.

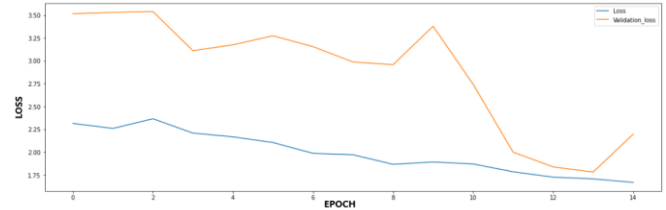
Fig. 18. Loss Graph for the CNN-RNN-LSTM Model TTSL Dataset

These results were better than the ASL Dataset but still a poor performance overall.

*D. CNN-TD (Time Distributed)*

*a) ASL Dataset:* The main result of the classification report is shown on Table XI. "Bye" and "Hello" had the best recall values.

TABLE XI. CLASSIFICATION REPORT PERCENTAGES BY SIGN FOR THE CNN TIME DISTRIBUTED MODEL AND ASL DATASET

| "Sign" | Precision | Recall | F1-Score |
|---|---|---|---|
| Bye | 100% | 100% | 100% |
| Happy | 100% | 33% | 50% |
| Hello | 18% | 100% | 31% |
| Help | 0% | 0% | 0% |
| How are you | 100% | 60% | 75% |
| Thanks | 100% | 50% | 67% |

The accuracy graph was not a smooth ascension with a significant dip between the 11th and 12th epoch for the validation accuracy, see Figure 18.

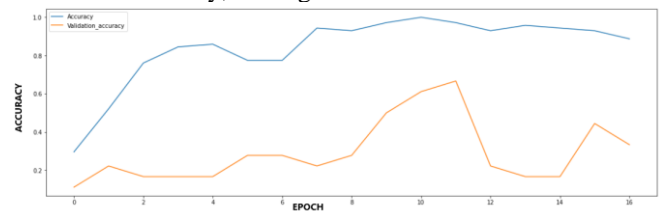
Fig. 19. Accuracy Graph for the CNN Time Distributed Model and ASL Dataset

The Loss graph similarly does not have a smooth descent as well with an uptick in the validation loss 12th and 14th epoch, see Figure 20.

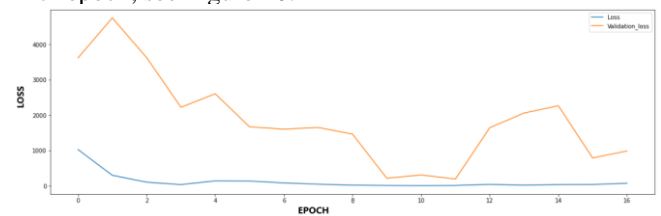
Fig. 20. Loss Graph for the CNN Time Distributed Model and ASL Dataset

Overall other than the anomalies the accuracy values outweighed the errors at each epoch. The large errors match the missing figures for one of the signs. There are some symptoms of overfitting as indicated by the extremely high loss figures compared to the accuracy at certain epochs. Time distribution wrapping makes a CNN more suited for temporal datasets, however, it was unexpected for it outperforms the other models such as CNN-RNN-LSTM.

*b) TTSL Dataset:*

The main result of the classification report is shown on Table XII. "Punch-a-Creme" again had the lowest recall value. This sign being one of three that had the least amount videos in the training dataset.

TABLE XII. CLASSIFICATION REPORT PERCENTAGES BY SIGN FOR THE CNN TIME DISTRIBUTED MODEL AND TTSL DATASET

| "Sign" | Precision | Recall | F1-Score |
|---|---|---|---|
| Doubles | 100% | 100% | 100% |
| Parang | 80% | 100% | 89% |
| Punch-a-Creme | 100% | 50% | 67% |
| Sorrel | 100% | 100% | 100% |

The Accuracy graph was completely flat with high values, see Figure 21.

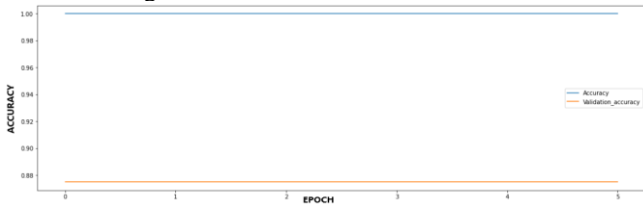

Fig. 21. Accuracy Graph for the CNN Time Distributed Model and TTSL Dataset

The Loss graph similarly was completely flat with high validation loss values again suggesting some overfitting during training, see Figure 22.

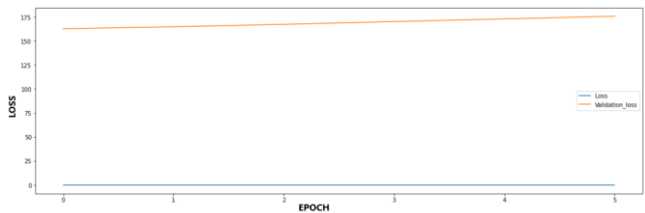

Fig. 22. Loss Graph for the CNN Time Distributed Model and TTSL Dataset

*E. Assessing a new Sign*

It was shown that 3DCNN and CNN-TD were the best models to identify TTSL with noted better percentages for accuracy and for recalling signs. Both models having 87.5% recall, see Section IV, subsections A-D. In addition, 3DCNN showed fewer resource requirements, see Section IV, Table III. With these results, these two models were identified for testing a simulation in assessing a new sign. The first step was saving the models identified that were trained with the TTSL dataset.

This simulation was done in Google Colab by saving the model for reuse after training. This was done with the function model.save() to be loaded later using model.load(). The video was recorded by accessing the webcam and saving the result to google drive. The video pre-processing and the prediction assessment was run on the video created and the results printed. This experiment was assessed based on the ability of the model to translate the sign correctly as well as the time taken for pre-processing and prediction. The certainty of the prediction was also printed, to double as a grade for the signer's "form", see Figure 23 for snippets of code from the test.

Fig. 23. Code snippets from Google Colab depicting: (1) saving the model, (2) loading the model for the experiment, (3) recording the video and (4) the translation code.

On the first execution of the translation process the results were displayed after one minute for the CNN-TD model and after 10 seconds for the 3DCNN model. However, when the subsequent signs provided by the user were tested, the predictions were displayed much faster; after 3 to 10 seconds.

All the TTSL signs mentioned in this project were tested and the ideal model 3DCNN, performed much better that the CNN-TD Model with the CNN-TD model only able to correctly identify the sign Doubles. All TTSL signs tested on the saved CNN-TD model in this experiment were translated incorrectly with 100% certainty on Doubles. Table XIII presents the results including all percentages summarized from the experiment. This was an interesting result as the recall percentages were high for the TTSL dataset the classification report. The Doubles dataset being the largest of the TTSL dataset may have skewed results for accurate prediction of new data.

TABLE XIII. RESULTS OF 'REAL-TIME'/NEW SIGN VIDEO TRANSLATION ON GOOGLE COLAB

| Sign | CNN-TD (T/F) | CNN-TD (2nd choice) | 3DCNN (T/F) | 3DCNN (2nd choice) |
|---|---|---|---|---|
| Doubles | T -Doubles 100% 10 Seconds | N/A | T-Doubles 99% 3 Seconds | F-Sorrel 79% |
| Parang | F -Doubles 100% 5 Seconds | N/A | T- Parang 98% 3 Seconds | F-Sorrel 56% |
| Punch -aCreme | F -Doubles 100% 5 Seconds | N/A | T-Punch-a-creme 93% 7 Seconds | F-Sorrel 60% |
| Sorrel | F -Doubles 100% 60 Seconds | N/A | F-Punch-a-creme 78% 3 Seconds | T-Sorrel 65% |

Whereas the 3DCNN trained Model was able to correctly identify 3 out of 4 signs. The only incorrectly translated sign was Sorrel, it was identified as "Punch-a-creme" with 78% certainty. The second highest prediction for the sign attempt was Sorrel with a prediction percentage of 65%. A possible challenge identified was the dataset being extremely similar in terms of the signs being completed by the same person in the same environment. The signs had to be executed as close to the original dataset or risk incorrect translation completely. While for a teaching system the goal is for the student to copy the form of the teacher, the attempted sign should have some more room for dissimilarities. A variation in persons executing the signs and a varied environment would benefit this 3DCNN model for further training, for it to be reimplemented for a live system.

## V. Discussion

Dataset size had the greatest impact on results, second to this was the model setup for certain models. Best in terms of accuracy and overall results for both TTSL and ASL was the 3DCNN approach followed by the CNN—LSTM models. The CNN-TD was the surprise contender although it must be noted that during training there was some overfitting. Considering TTSL only, the top performers are 3DCNN followed by CNN-TD. The high variances in Accuracy and Loss in the CNN-RNN-LSTM indicate that the setup of the model was too complex for the dataset being used. Some adjustment and improvement in model setup are required to improve results. Resource usage was also noted as higher on this model, see Section IV, Table III. Comparing the results of the experiments in Section IV, Subsections A-D the best performing model overall is the 3DCNN model. Not too far behind was the CNN-LSTM (ConvnetLSTM2D) and CNN-TD models.

ConvLSTM2D (CNN-LSTM) performed better than the CNN and RNN's LSTM combinations. The transfer learning utilized in the CNN-RNN-LSTM method resulted in lost or untrainable parameters that none of the other more successful models did experience. CNN-TD was a surprisingly high performer. Research in [28] shows that all the other investigated models outperformed CNN-TD stand-alone for video spatiotemporal datasets. While this remained true for the ConvLSTM2D model application, the time distributed execution had a better result overall than CNN-RNN-LSTM model.

The assessing of a new sign experiment showed promise for future work. It re-emphasized how ideal the 3DCNN model is for TTSL Translation versus CNN-TD. This was done on the two best performing models for TTSL which were 3DCNN followed by CNN-TD. The results showed the superiority of the 3DCNN Model to correctly identify signs. In addition, 3DCNN was faster than the CNN-TD in terms of execution time and more accurate in terms of newly submitted signs to translate. However, more training needs to be done on a larger, more varied dataset.

Despite the warning on the 3DCNN model's tendency to drain on run time and processing resources because of large parameter requirements, this model RAM and GPU usage was the lowest compared to the rest of the models. It did have the highest parameter values as predicted. It is, however, worth training a larger dataset on this model as it could only improve the results. This model also outperformed the others mentioned in terms of accuracy and would be a good base model for future work in this direction.

## VI. Conclusion

Signs are dynamic gestures characterized by continuous hand motions and hand configurations [5]. A neural network approach was chosen as the method to read, understand and translate sign language, NNs are known to adjust to changing input, as such the network produces the best possible outcome without needing to revamp the output criteria. Once paired with adequate tools NNs can have faster processing on large complex datasets. This paper explores a Trinidad and Tobago sign language dataset, recognition of the signs by famous NN models, with the experimentation done identifying 3DCNN as the best overall performing model in terms of accuracy and efficiency, and finally a demonstration or simulation done of the assessment of a TTSL sign using the saved, trained 3CDNN model.

Based on the findings on identifying signs, it is possible to develop a working system or application that translates and corrects meaningful sign language into audible or written words for others to understand, using the 3DCNN Model as the foundation. To increase the appreciation of the performance of this model reporting can be added on the level accuracy without a model. Adding the ability to acquire real-time confirmation and correction of signs means it can evolve into a supplemental form of education. It can be utilized by sign language students to translate or grade practice signs and reinforce what is learned in sign language classroom sessions by either correcting or confirming correct sign language forms. This application can also be expanded to teach Trinidad and Tobago Sign language and perhaps other Caribbean Sign Languages in schools regionwide. Overall, there is still much work that can be done regarding a Sign Language Recognition system for TTSL and Caribbean SL. Hopefully, a tutorial or education application can be implemented in a real-life scenario soon.